\relax
\documentclass[letterpaper]{article}

\usepackage{aaai20}
\usepackage{times}  
\usepackage{helvet}  
\usepackage{courier}  
\usepackage{url}  
\usepackage{graphicx}  
\usepackage{booktabs}       
\usepackage{amsfonts}       
\usepackage{nicefrac}       
\usepackage{microtype}      
\usepackage{graphicx}
\usepackage{amsmath,amsfonts,amsthm,bm, amssymb}
\usepackage{caption}
\usepackage{subcaption}
\usepackage{color}
\usepackage{gensymb}
\usepackage{url}
\usepackage{verbatim}
\usepackage{placeins}
\newcommand{\chris}[1]{}
\newcommand{\changhao}[1]{}
\newcommand{\andrew}[1]{}

\begin{document}
\title{AtLoc: Attention Guided Camera Localization}

\author{Bing Wang,
Changhao Chen,
Chris Xiaoxuan Lu,
Peijun Zhao,
\\
\Large{\textbf{Niki Trigoni,
Andrew Markham
}}
\\
{Department of Computer Science, University of Oxford}\\
firstname.lastname@cs.ox.ac.uk
}

\maketitle
\begin{abstract}
Deep learning has achieved impressive results in camera localization, but current single-image techniques typically suffer from a lack of robustness, leading to large outliers. To some extent, this has been tackled by sequential (multi-images) or geometry constraint approaches, which can learn to reject dynamic objects and illumination conditions to achieve better performance. In this work, we show that attention can be used to force the network to focus on more geometrically robust objects and features, achieving state-of-the-art performance in common benchmark, even if using only a single image as input. Extensive experimental evidence is provided through public indoor and outdoor datasets. Through visualization of the saliency maps, we demonstrate how the network learns to reject dynamic objects, yielding superior global camera pose regression performance. The source code is avaliable at \url{https://github.com/BingCS/AtLoc}.
\end{abstract}
\section{Introduction}
\andrew{high level comments: It is clear that this technique works better than SOTA, even better than sequential, which is in itself quite amazing. But I don't see any deep insight into either why it works better or why it is not a trivial addition (i.e. just add attention to any network and it inevitably gets better). I unfortunately don't have a good suggestion on how to make this clearer. Intuitively, relocalization fails when images which are spatially far from one another have similar features in the embedding space. Can you demonstrate or show that attention makes images "cluster" better that are spatially similar? Perhaps an information theoretic metric, or t-SNE embedding, or some insight into the manifold would help understand why attention is better at "hashing" distant images to different embeddings}

Location information is of key importance to wide variety of applications, from virtual reality to delivery drones, to autonomous driving.
One particularly promising research direction is camera pose regression or localization - the problem of recovering the 3D position and orientation of a camera from an image or set of images. 



Camera localization has been previously tackled by exploiting the appearance and geometry in a 3D scene, for example, key points and lines, but suffers from performance degradation when deployed in the wild~\cite{brachmann2017dsac,walch2017image}. This is due to the fact that the handcrafted features change significantly across different scenarios due to lighting, blur and scene dynamics leading to poor global matches. Recent deep learning based approaches are able to automatically extract features and directly recover the absolute camera pose from a single image, without any hand-engineering effort, as was demonstrated in the seminal PoseNet~\cite{kendall2015posenet}. Extensions include the use of different encoder networks e.g. ResNet in PoseNet Hourglass~\cite{melekhov2017image} or geometric constraints~\cite{kendall2017geometric}. Although these techniques show good performance in general, they are plagued by a lack of robustness when faced with dynamic objects or changes in illumination. This is particularly apparent in outdoor datasets where scenes are highly variable e.g. due to moving vehicles or pedestrians.

\begin{figure}
    \centering
    \includegraphics[width=1\linewidth]{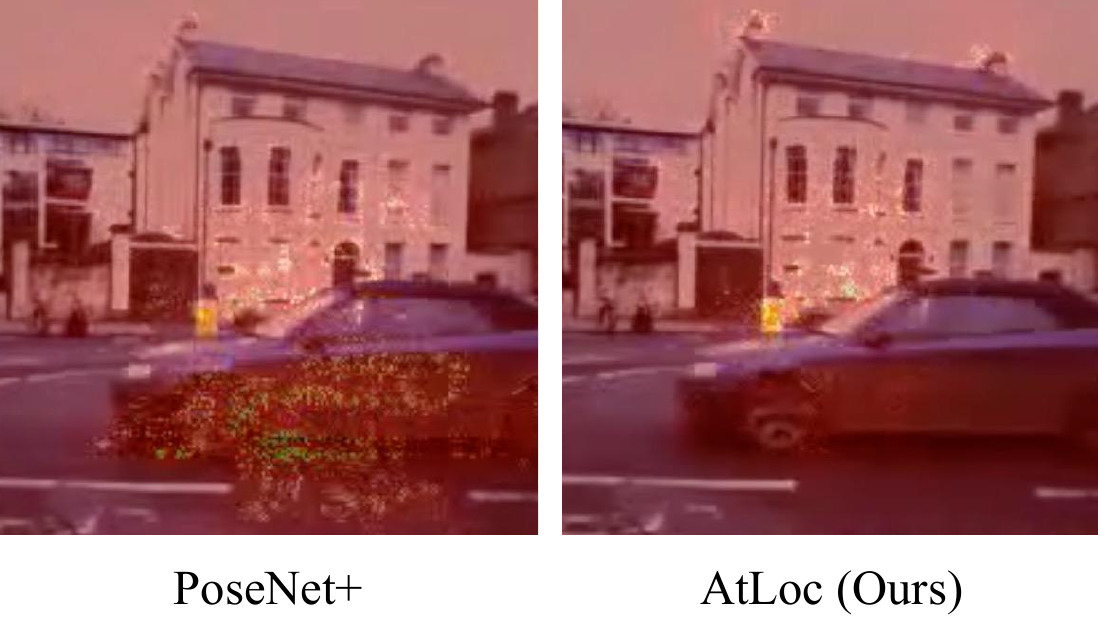}
    \vspace{-0.5cm}
    \caption{Saliency maps of one scene selected from Oxford RobotCar~\cite{maddern20171} indicate that AtLoc is able to force the neural network model to focus on geometrically robust objects (e.g. building structures in the right) rather than environmental dynamics (e.g. moving vehicles in the left) compared with PoseNet+~\cite{brahmbhatt2018geometry}. }
    \label{fig:opening}
\vspace{-0.5cm}
\end{figure}
To tackle this lack of robustness, further techniques have considered using multiple images as input to the network, with the premise being that the network can learn to reject temporally inconsistent features across frames. Examples include VidLoc~\cite{clark2017vidloc} and the recent MapNet~\cite{brahmbhatt2018geometry} which achieves state-of-the-art performance in camera pose regression. 

 \begin{figure*}[t]
        \centering
        \includegraphics[width=1.0\linewidth]{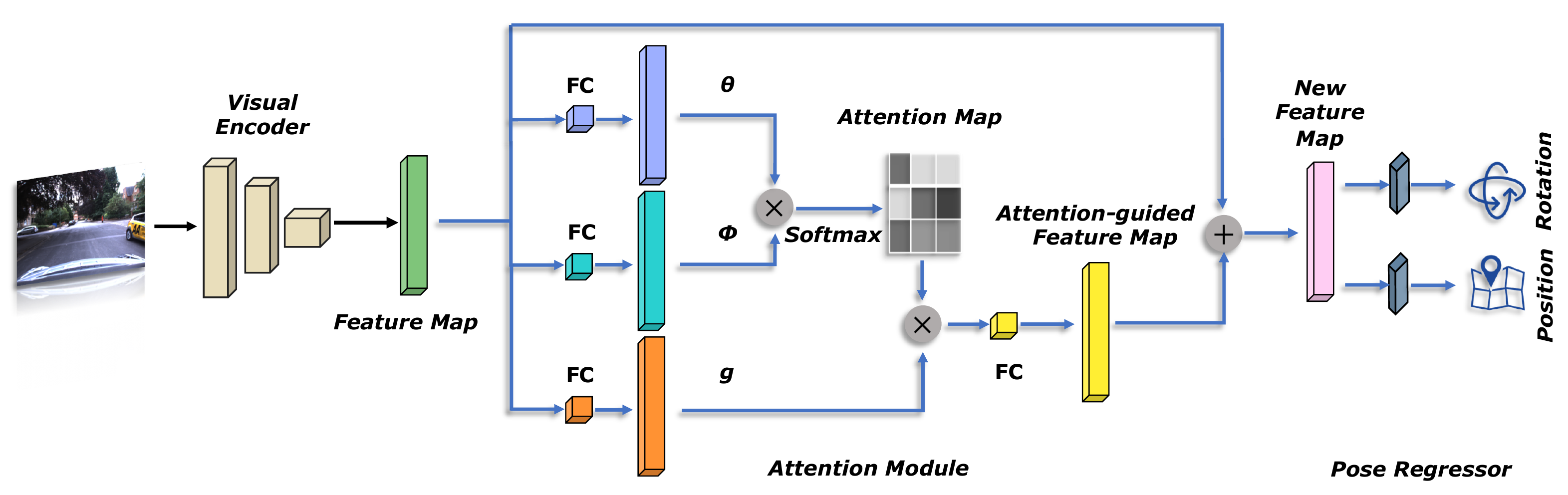}
        \vspace{-0.5cm}
            \caption{An overview of our proposed AtLoc framework, consisting of Visual Encoder (extracts features from a single image), Attention Module (computes the attention and reweights the features), and Pose Regressor (maps the new features into the camera pose).}    
        \label{fig:framework}
        \vspace{-0.5cm}
\end{figure*}

In this work we pursue an alternative approach to achieve robust camera localization and ask if we can achieve or even surpass the performance of multi-frame, sequential techniques by learning to attentively focus on parts of the image that are temporally consistent and informative e.g. buildings, whilst ignoring dynamic parts like vehicles and pedestrians, using a single image as input, as shown in Figure \ref{fig:opening}. We propose \textbf{AtLoc} - \textbf{At}tention Guided Camera \textbf{Loc}alization, an attention based pose regression framework to recover camera pose. Unlike previous methods, our proposed AtLoc does not require sequential (multiple) frame nor geometry constraints designed and enforced by humans. 

We show that our model outperforms previous techniques, and achieves state-of-the-art results in common benchmarks. It works efficiently across both indoor and outdoor scenarios and is simple and end-to-end trainable without requiring any hand-crafted geometric loss functions. We provide detailed insight into how incorporating attention allows the network to achieve accurate and robust camera localization.

The main contributions of this work are as follows:
\begin{itemize}
    \item We propose a novel self-attention guided neural network for single image camera localization, allowing accurate and robust camera pose estimation.
    \item By visualizing the feature salience map after the attention, we show how our attention mechanism encourages the framework to learn stable features.
    \item Through extensive experiments in both indoor and outdoor scenarios, we show that our model achieves state-of-the-art performance in pose regression, even outperforming multiple frame (sequential) methods.
\end{itemize}

\section{Related Work}
\paragraph{Deep Neural Networks for Camera Localization}
Recent attempts have investigated the camera localization using deep neural networks (DNNs). Compared with traditional structure-based methods \cite{chen2011city,liu2017efficient} and image retrieval-based methods \cite{sattler2012improving,arandjelovic2016netvlad}, DNN-based camera localization methods can automatically learn features from data rather than building a map or a database of landmark features by hand \cite{sattler2019understanding}.
As the seminal work in this vein, PoseNet\cite{kendall2015posenet} is the first one to adopt deep neural network to estimate camera pose from a single image. This approach is then extended by leveraging RNNs (e.g. LSTM) to spatially \cite{walch2017image,wang2018end} and temporally \cite{clark2017vidloc} improve localization accuracy. Later on, localization performance is further improved by estimating uncertainty of the global camera pose with Bayesian CNN \cite{kendall2016modelling,cai2018hybrid} and replacing feature extraction architecture with Residual Neural Network \cite{melekhov2017image}. 
However, the aforementioned approaches rely on the hand-tuned scale factor to balance the position and rotation losses during learning process. 
To address this issue, a learning weighted loss and a geometric reprojection loss \cite{kendall2017geometric} are introduced to produce more precise results. Recent efforts additionally leverage the geometric constraints from paired images \cite{brahmbhatt2018geometry,huang2019prior}, augment training data by synthetic generation\cite{purkait2018synthetic} or introduce pose-graph optimization with neural graph model\cite{parisotto2018global}. Instead of imposing the temporal information or geometry constraints as the previous work, we developed an attention mechanism for DNN-based camera localization to self-regulate itself, and automatically learn to constrain the DNNs to focus on geometrically robust features. Our model outperforms previous approaches, and achieves state-of-the-art results in common benchmarks.
\paragraph{Attention Mechanism}
Our work is related with the self-attention mechanisms, which have been widely embedded in various models capturing long-term dependencies \cite{bahdanau2014neural,xu2015show,yang2019context}. 
Self-attention was initially designed for machine translation \cite{vaswani2017attention,dou2019dynamic,cheng2016long}, achieving the state-of-the-art performance. It is also integrated with an autoregressive model to generate image as Image Transformer \cite{parmar2018image,kingma2018glow}. Another usage is to be formalized as a non-local operation to capture the spatial-temporal dependencies in video sequences \cite{wang2018non,yuan2019find}. 
A similar non-local architecture was introduced to Generative Adversarial Networks (GANs) for extracting global long-range dependencies \cite{zhang2018self,liu2019perceptual}. 
\cite{parisotto2018global} used an attention-based recurrent neural network for back-end optimization in a SLAM system, but not for camera relocalization.
Despite its successes in a wide range of of computer vision \cite{fu2019sta,chen2019selective} and natural language process tasks, self-attention has never been explored in camera pose regression. 
Our work integrated non-local style self-attention mechanism into the camera localization model to show the effectiveness of correlating robust key features and improve model performance.

\section{Attention Guided Camera Localization}
This section introduces Attention Guided Camera Localization (AtLoc), an self-attention based deep neural network architecture to learn camera poses from a single image. 
Figure~\ref{fig:framework} illustrates a modular overview of the proposed framework, consisting of a visual encoder, an attention module and a pose regressor. 
The scene of a single image is compressed into an implicit representation by the visual encoder.
Conditioned on the extracted features, the attention module computes the self-attention maps to re-weight the representation into a new feature space.
The pose regressor further maps the new features after the attention operators into the camera pose, i.e. the 3-dimensional location and 4-dimensional quaternion (orientation).

\subsection{Visual Encoder}
The visual encoder serves to extract features that are necessary for the pose regression task, from a single monocular image.
Previous works\cite{kendall2017geometric,brahmbhatt2018geometry} showed successful applications of the classical convolutional neural network (CNN) architectures in camera pose estimation, e.g. GoogleNet \cite{szegedy2015going} and ResNet \cite{he2016deep}. 
Among them, the ResNet based \cite{brahmbhatt2018geometry} frameworks achieved more stable and precise localization results over other architectures, due to the fact that the residual networks allow to train deeper layers of neural networks and reduce the gradient vanishing problems. Therefore, we considered to adopt a residual network with 34 layers (ResNet34) as the foundation for the visual encoder in the proposed AtLoc model.
Here, the weights of ResNet34 were initialized with the ResNet34 pretrained with the image classification on ImageNet dataset \cite{imagenet}. 
To encourage learning meaningful features for a pose regression, the base network is further modified by replacing the final 1000 dimensional fully-connected layer with a $C$ dimensional fully-connected layer and removing the Softmax layers used for classification. $C$ is the dimension of the output feature. Considering the efficiency and performance of the model, the dimension is chosen as $C=2048$. Given an image $\mathbf{I} \in \mathbb{R}^{\hat{C} \times H \times W}$, the features $\mathbf{x} \in \mathbb{R}^{C}$ can be extracted via the visual encoder $f_{\text{encoder}}$:
\begin{equation}
    \mathbf{x} = f_{\text{encoder}} (\mathbf{I})
\end{equation}


\subsection{Attention Module}
Although the ResNet34 based visual encoder is capable of automatically learning the necessary features for camera localization, the neural network trained in certain specific scenes can be overfitted into the featureless appearance or the environmental dynamics. This will impact the generalization capacity of the model, and degrade the model performance in testing sets, especially in the outdoor scenarios due to the moving vehicles or weather change. Unlike the previous trials by introducing the temporal information \cite{clark2017vidloc} or geometric constraints \cite{brahmbhatt2018geometry}, we propose to adapt a self-attention mechanism into our framework. As the Figure \ref{fig:framework} shown, this self-attention module is conditioned the features extracted by the visual encoder, and generates an attention map to enforce the model to focus on stable and geometry meaningful features. 
It is able to self-regulate itself without any hand-engineering geometry

s or prior information.

We adopt a non-local style self-attention, which has been applied in video analysis \cite{wang2018non} and image generation \cite{zhang2018self}, in our attention module. This aims to capture the long-range dependencies and global correlations of the image features, which will help generate better attention-guided feature maps from widely separated spatial regions \cite{wang2018non}.



The features $\mathbf{x} \in \mathbb{R}^{C}$ extracted by the visual encoder are first used to compute the dot-product similarity between two embedding spaces $\theta(\mathbf{x}_i)$ and $\phi(\mathbf{x}_j)$:
    \begin{equation}
	    S(\mathbf{x}_i, \mathbf{x}_j) = \theta(\mathbf{x}_i)^T\phi(\mathbf{x}_j),
	    \label{eq:similarity}
    \end{equation}
where embeddings $\theta(\mathbf{x}_i)=\mathbf{W}_\theta \mathbf{x}_i$ and $\phi(\mathbf{x}_j)=\mathbf{W}_\phi \mathbf{x}_j$ linearly transform features at the position $i$ and $j$ into two feature spaces respectively. 

The normalization factor $C$ is defined as the $C(\mathbf{x}_i)= \sum_{\forall j} S(\mathbf{x}_i, \mathbf{x}_j)$ with all feature position $j$.
Given another linear transformation $g(\mathbf{x}_j)=\mathbf{W}_g \mathbf{x}_j$, the output attention vector $\mathbf{y}$ is calculated via:
    \begin{equation}
        \mathbf{y}_i = \frac{1}{C(\mathbf{x}_i)} \sum_{\forall j} S(\mathbf{x}_i, \mathbf{x}_j) g(\mathbf{x}_j) ,
    \end{equation}
where the attention vector $\mathbf{y}_i$ indicates to what extent the neural model focuses on the features $\mathbf{x}_i$ at the position $i$. Finally, the self-attention of input features $\mathbf{x}$ can be written as: 
\begin{equation}
	\mathbf{y} = \text{Softmax}(\mathbf{x}^T \mathbf{W}^T_\theta \mathbf{W}_\phi \mathbf{x}) \mathbf{W}_g \mathbf{x}
	\label{eq:attentionmap}
\end{equation}
Furthermore, we add a residual connection back to a linear embedding of the self-attention vectors:
\begin{equation}
	\text{Att}(\mathbf{x}) = \alpha(\mathbf{y}) + \mathbf{x},
	\label{eq:attoutput}
\end{equation}
where the linear embedding $\alpha(\mathbf{y})=\mathbf{W}_\alpha \mathbf{y}$ outputs a scaled self-attention vectors with learnable weights $\mathbf{W}_\alpha$.

In our proposed model, fully-connected layers are implemented to generate learned weight matrices $\mathbf{W}_\theta$, $\mathbf{W}_\phi$, $\mathbf{W}_g$ and $\mathbf{W}_\alpha$ in space $(\nicefrac{C}{n})$, where $C$ is the number of channels of the input feature $x$ and $n$ is the downsampling ratio for the attention maps. Based on extensive experiments, we found that $n=8$ performs best across different datasets.

\subsection{Learning Camera Pose}
\begin{table*}[t]
    \centering
    \tiny
    \resizebox{\linewidth}{!}{
    \begin{tabular}{l|cccccc}
         & PoseNet & Bayesian & PoseNet & Hourglass & PoseNet17 & \textbf{AtLoc}\\
        Scene & & PoseNet & Spatial LSTM & & & \textbf{(Ours)} \\
        \hline \hline
        Chess 		& 0.32m, 6.60\degree & 0.37m, 7.24\degree & 0.24m, 5.77\degree & 0.15m, 6.17\degree & 0.13m, 4.48\degree & \textbf{0.10m}, \textbf{4.07\degree} \\
        Fire 		& 0.47m, 14.0\degree & 0.43m, 13.7\degree & 0.34m, 11.9\degree & 0.27m, 10.8\degree & 0.27m, 11.3\degree & \textbf{0.25m}, \textbf{11.4\degree} \\
        Heads 		& 0.30m, 12.2\degree & 0.31m, 12.0\degree & 0.21m, 13.7\degree & 0.19m, 11.6\degree & 0.17m, 13.0\degree & \textbf{0.16m}, \textbf{11.8\degree} \\
        Office 		& 0.48m, 7.24\degree & 0.48m, 8.04\degree & 0.30m, 8.08\degree & 0.21m, 8.48\degree & 0.19m, 5.55\degree & \textbf{0.17m}, \textbf{5.34\degree} \\
        Pumpkin 	& 0.49m, 8.12\degree & 0.61m, 7.08\degree & 0.33m, 7.00\degree & 0.25m, 7.01\degree & 0.26m, 4.75\degree & \textbf{0.21m}, \textbf{4.37\degree} \\
        Kitchen & 0.58m, 8.34\degree & 0.58m, 7.54\degree & 0.37m, 8.83\degree & 0.27m, 10.2\degree & 0.23m, \textbf{5.35\degree} & \textbf{0.23m}, 5.42\degree \\
        Stairs 		& 0.48m, 13.1\degree & 0.48m, 13.1\degree & 0.40m, 13.7\degree & 0.29m, 12.5\degree & 0.35m, 12.4\degree & \textbf{0.26m}, \textbf{10.5\degree} \\
        \hline
        Average & 0.45m, 9.94\degree & 0.47m, 9.81\degree & 0.31m, 9.85\degree & 0.23m, 9.53\degree & 0.23m, 8.12\degree & \textbf{0.20m}, \textbf{7.56\degree} \\
    \end{tabular}
    }
    \caption{\textbf{Camera localization results on 7 Scenes (Without temporal s).} For each scene, we compute the median errors in both position and rotation of various single-image based baselines and our proposed method.}
	\label{tbl:7scenes}
	    \vspace{-0.5cm}
\end{table*}

\begin{table}[t]
	\centering
	\caption{\textbf{Training and testing Sequences of Oxford RobotCar.} LOOP is a relatively shorter subset ($1120m$ in total length) and FULL covers a length of $9562m$. 
	}
			\begin{tabular}{l|ccc}
				Sequence & Time & Tag &  Mode \\
				\hline
				\hline
				--    & 2014-06-26-08-53-56 & overcast & Training \\
				--    & 2014-06-26-09-24-58 & overcast & Training \\
				LOOP1 & 2014-06-23-15-41-25 & sunny    & Testing  \\
				LOOP2 & 2014-06-23-15-36-04 & sunny    & Testing  \\
				\hline
				--    & 2014-11-28-12-07-13 & overcast & Training \\
				--    & 2014-12-02-15-30-08 & overcast & Training \\
				FULL1 & 2014-12-09-13-21-02 & overcast & Testing  \\
				FULL2 & 2014-12-12-10-45-15 & overcast & Testing  \\
			\end{tabular}
    \vspace{-0.5cm}
	\label{tab:train_test_split}
\end{table}

The pose regressor maps the attention guided features $\text{Att}(\mathbf{x})$ to location $\mathbf{p} \in \mathbb{R}^3 $ and quaternion $\mathbf{q} \in \mathbb{R}^4 $ respectively through Multilayer Perceptrons (MLPs): 
    \begin{equation}
        [\mathbf{p}, \mathbf{q}] = \text{MLPs}(\text{Att}(\mathbf{x}))
    \end{equation}
Given training images $\mathbf{I}$ and their corresponding pose labels $\mathbf{[\hat{p}, \hat{q}]}$ represented by the camera position $\mathbf{\hat{p}} \in \mathbb{R}^3 $ and a unit quaternion $\mathbf{\hat{q}} \in \mathbb{R}^4 $ for orientation, the parameters inside the neural networks are optimized with L1 Loss via the following loss function \cite{brahmbhatt2018geometry}:
\begin{equation}
	loss(\mathbf{I}) =\Vert \mathbf{p}-\mathbf{\hat{p}} \Vert_{1} e^{-\beta} + \beta +
    \Vert \mathbf{\log q}-\mathbf{\log \hat{q}} \Vert_{1} e^{-\gamma} + \gamma
	\label{eq:loss}
\end{equation}
where $\beta$ and $\gamma$ are the weights that balance the position loss and rotation loss.
$\mathbf{\log q}$ is the logarithmic form of an unit quaternion $\mathbf{q}$, which is defined as:
\begin{equation}
    \log\mathbf{q}= \begin{cases}
            \frac{\mathbf{v}}{\Vert\mathbf{v}\Vert}\cos^{-1}u,& \text{if } \Vert\mathbf{v}\Vert \neq 0\\
            \mathbf{0},& \text{otherwise}
        \end{cases}
\end{equation}
Here, $\mathbf{u}$ denotes the real part of an unit quaternion while $\mathbf{v}$ is its imaginary part. For all scenes, both $\beta$ and $\gamma$ are simultaneously learned during training with approximate initial values of $\beta_0$ and $\gamma_0$. In camera pose regression tasks, quaternions are widely used to represent the orientation due their ease  of formulation in a continuous and differentiable way. By normalizing any 4D quaternions to unit length, we can easily map any rotations in 3D space to valid unit quaternions. But this has one main issue: quaternions are not unique. In practice, both $\mathbf{-q}$ and $\mathbf{q}$ can represent the same rotation because a single rotation can be mapped to two hemispheres. To ensure that each rotation only has a unique value, all quaternions are restricted to the same hemisphere in this paper.

\subsection{Temporal Constraints}
Sharing the same flavor with geometry-aware learning methods \cite{brahmbhatt2018geometry,xue2019local,huang2019prior}, we extend our proposed \textbf{AtLoc} to \textbf{AtLoc+} by incorporating temporal constraints between image pairs. Intuitively, temporal constraints can enforce the learning of globally consist features, and thereby improve the overall localization accuracy. In this work, the loss considering temporal constraints is defined as:
\begin{equation}
	loss(\mathbf{I}_{total}) =loss(\mathbf{I}_i) + \alpha \sum_{i\neq j} loss(\mathbf{I}_{ij})
	\label{eq:loss_seq}
\end{equation}
where $\mathbf{i}$ and $\mathbf{j}$ indicate the index of images. $\mathbf{I}_{ij} = (\mathbf{p}_i\!-\!\mathbf{p}_j, \mathbf{q}_i\!-\!\mathbf{q}_j)$ represents the relative pose between images $\mathbf{I}_i$ and $\mathbf{I}_j$. $\mathbf{\alpha}$ denotes the weight coefficient between the loss of the absolute pose from a single image and the relative pose from image pairs.

\begin{table}[t]
    \centering
        \caption{\textbf{Camera localization results on 7 Scenes (With temporal Constraints).} For each scene, we compare the median errors in both position and rotation of VidLoc, MapNet and our approach.}
    \resizebox{\linewidth}{!}{
    \begin{tabular}{l|cccccc}
        & VidLoc & MapNet & \textbf{AtLoc+}\\
        Scene & & & \textbf{(Ours)}\\
        \hline \hline
        Chess 		& 0.18m, NA & \textbf{0.08m}, 3.25\degree & 0.10m, \textbf{3.18}\degree  \\
        Fire 		& 0.26m, NA & 0.27m, 11.7\degree & \textbf{0.26m}, \textbf{10.8}\degree  \\
        Heads 		& 0.14m, NA & 0.18m, 13.3\degree & \textbf{0.14m}, \textbf{11.4}\degree  \\
        Office 		& 0.26m, NA & 0.17m, \textbf{5.15}\degree & \textbf{0.17m}, 5.16\degree  \\
        Pumpkin 	& 0.36m, NA & 0.22m, 4.02\degree & \textbf{0.20m}, \textbf{3.94}\degree  \\
        Kitchen & 0.31m, NA & 0.23m, 4.93\degree & \textbf{0.16m}, \textbf{4.90}\degree  \\
        Stairs 		& 0.26m, NA & 0.30m, 12.1\degree & \textbf{0.29m}, \textbf{10.2}\degree  \\
        \hline
        Average & 0.25m, NA & 0.21m, 7.77\degree & \textbf{0.19m}, \textbf{7.08\degree}  \\
    \end{tabular}
    }
	\label{tbl:7scenes_Seq}
	        \vspace{-0.5cm}
\end{table}

\section{Experiments}
\begin{figure}[t]
    \centering
    \includegraphics[width=0.85\linewidth]{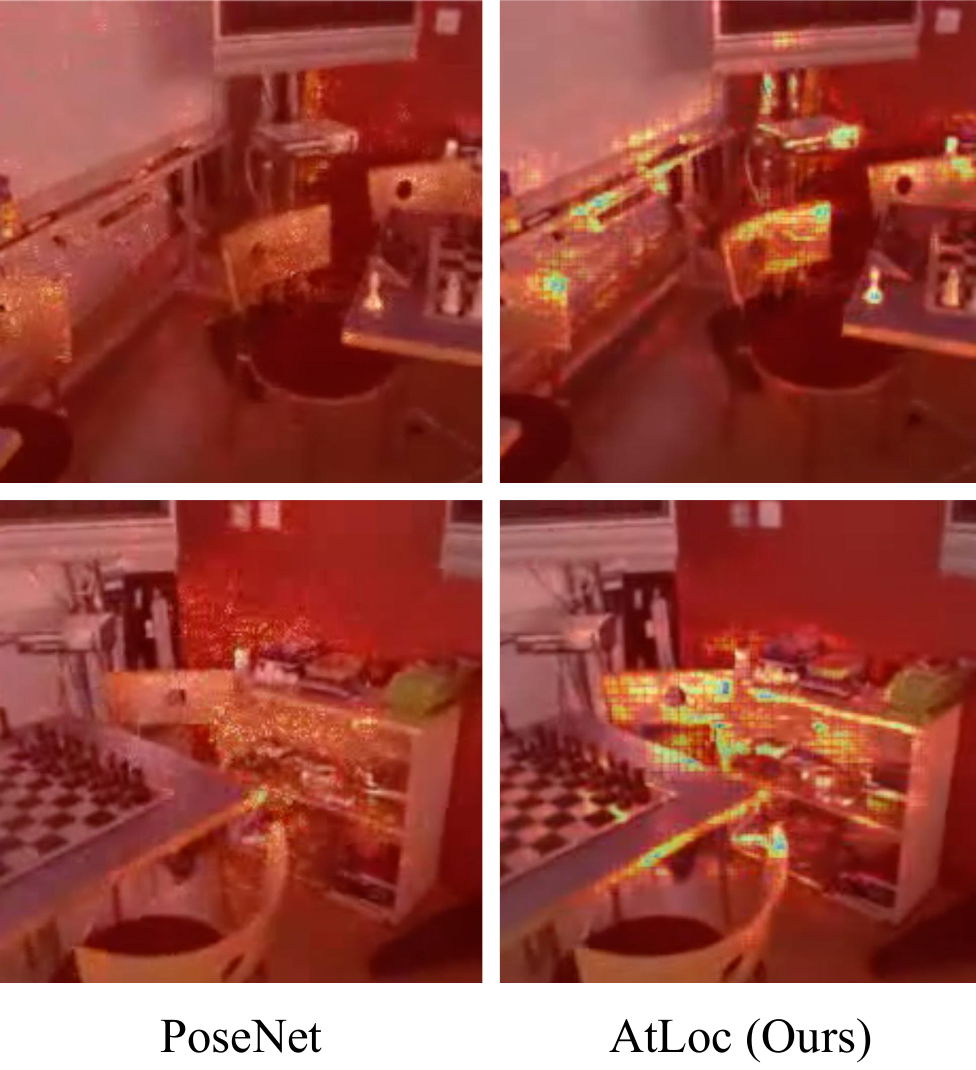}
    \vspace{-0.2cm}
    \caption{\textbf{Saliency maps of two scenes selected from Chess.} Each scene contains the saliency maps generated by PoseNet (left) and AtLoc (right) using attention.}
    \label{fig:chess_att}
\vspace{-0.5cm}
\end{figure}

\begin{figure}[t]
    \centering
    \includegraphics[width=0.85\linewidth]{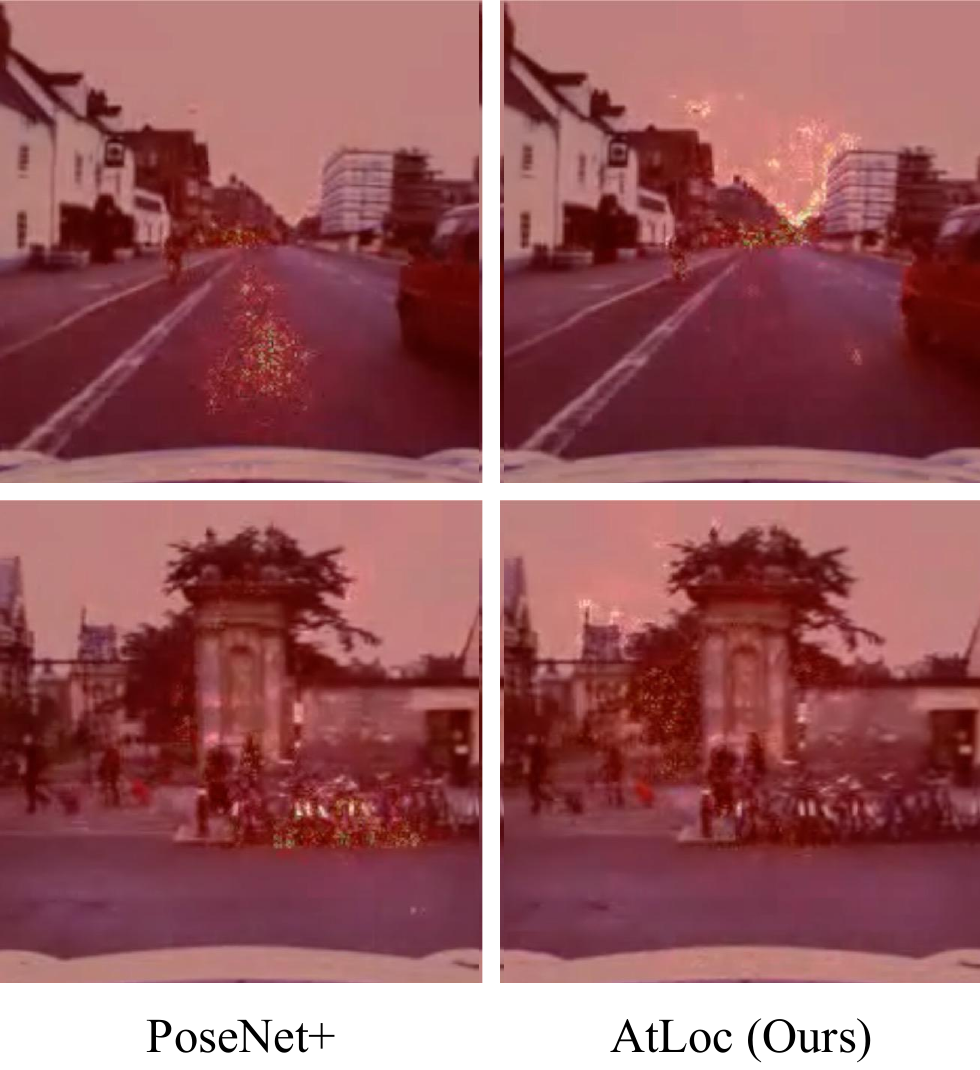}
     \vspace{-0.2cm}
    \caption{\textbf{Saliency maps of two scenes selected from Oxford RobotCar} generated from models without attention (left: PoseNet+) and with attention (right: AtLoc). Note how AtLoc learns to ignore visually uninformative features e.g. the road in the top figure and instead focus on more distinctive objects e.g. the skyline in the distance. AtLoc also learns to reject affordable objects e.g. the bicycles in the bottom figure, yielding more robust global localization.}
    \label{fig:robotcar_att}
\vspace{-0.5cm}
\end{figure}

\begin{table*}
	\centering
    \resizebox{\linewidth}{!}{
    \begin{tabular}{l|cc|cc|cc|cc}
               Sequence & \multicolumn{2}{c}{PoseNet+} & \multicolumn{2}{c|}{\textbf{AtLoc (Ours)}} & \multicolumn{2}{c}{MapNet} & \multicolumn{2}{c}{\textbf{AtLoc+ (Ours)}} \\
        \hline 
        \hline
        --          & Mean               &Median                 & Mean               & Median               & Mean             & Median               & Mean             & Median    \\
        LOOP1 		& 25.29m, 17.45\degree &  6.88m, \textbf{2.06}\degree & \textbf{8.61m}, \textbf{4.58}\degree & \textbf{5.68m}, 2.23\degree & 8.76m, \textbf{3.46\degree} & 5.79m, \textbf{1.54\degree} & \textbf{7.82m}, 3.62\degree & \textbf{4.34m}, 1.92\degree\\
        LOOP2 		& 28.81m, 19.62\degree &  5.80m, 2.05\degree & \textbf{8.86m}, \textbf{4.67}\degree & \textbf{5.05m}, \textbf{2.01}\degree & 9.84m, 3.96\degree &  4.91m, \textbf{1.67\degree} & \textbf{7.24m}, \textbf{3.60}\degree & \textbf{3.78m}, 2.04\degree \\
        FULL1 		& 125.6m, 27.10\degree & 107.6m, 22.5\degree & \textbf{29.6m}, \textbf{12.4\degree} & \textbf{11.1m}, \textbf{5.28\degree} & 41.4m, 12.5\degree & 17.94m, 6.68\degree & \textbf{21.0m}, \textbf{6.15}\degree & \textbf{6.40m}, \textbf{1.50}\degree \\
        FULL2 		& 131.1m, 26.05\degree & 101.8m, 20.1\degree & \textbf{48.2m}, \textbf{11.1\degree} & \textbf{12.2m}, \textbf{4.63\degree} & 59.3m, 14.8\degree & 20.04m, 6.39\degree & \textbf{42.6m}, \textbf{9.95}\degree & \textbf{7.00m}, \textbf{1.48}\degree \\
        \hline
        Average     & 77.70m, 22.56\degree & 55.52m, 11.7\degree & \textbf{23.8m}, \textbf{8.19\degree} & \textbf{8.54m}, \textbf{3.54\degree} & 29.8m, 8.68\degree & 12.17m, 4.07\degree & \textbf{19.7m}, \textbf{5.83}\degree & \textbf{5.38m}, \textbf{1.74}\degree\\
    \end{tabular}
    }
    \caption{\textbf{Camera localization results on the LOOP and FULL of the Oxford RobotCar.} For each sequence, we calculate the median and mean errors of position and rotation of Posenet+, MapNet and our approaches. Posenet and AtLoc leverage a single image while MapNet and AtLoc+ utilize sequential ones.}
\label{tbl:robotcar_results}
\vspace{-0.5cm}
\end{table*}
To train the proposed network consistently on different datasets, we rescale the images such that the shorter side is of length 256 pixels. The input images are then normalized to have  pixel intensities within the range -1 to 1. The ResNet34 \cite{he2016deep} component in our network is initialized by using a pretrained model on the ImageNet dataset while the remaining components follow random initialization. $256\times256$ pixels images are cropped for our network during both the training and testing phase with random and central cropping strategy respectively. For the training on Oxford RobotCar dataset, random ColorJitter is additionally applied when performing data augmentation, with values of 0.7 for brightness, contrast and saturation setting and 0.5 for hue. We note that this augmentation step is essential to improve the generalization ability of model over various weather and time-of-day conditions. We implement our approaches with PyTorch, using the ADAM solver \cite{kingma2014adam} and an initial learning rate of $5\times10^{-5}$. The network is trained on a NVIDIA Titan X GPU with the following hyperparameters: mini-batch size of 64, dropout rate probability of 0.5 and weight initializations of $\beta_0=0.0$ and $\gamma_0=-3.0$. When introducing temporal constraints, we sample consecutive triplets every 10 frames with $\alpha_0=1.0$ and initialize weight coefficient $\alpha_0=1.0$.

\subsection{Datasets and Baselines}

\paragraph{7 Scenes} \cite{shotton2013scene}
is a dataset consisting of RGB-D images from seven different indoor scenes captured by a handheld Kinect RGB-D camera. The corresponding ground truth camera poses were calculated using KinectFusion. All images were captured in a small-scale indoor office environment at the resolution of $640\times480$ pixels. Each scene contains two to seven sequences in a single room for training/testing, with 500 or 1000 images for each sequence. As a a popular dataset for visual relocalization, the sequences contained in this dataset were recorded under various camera motion status and different conditions, e.g. motion blur, perceptual aliasing and textureless features in the room.
\begin{figure*}[t]
    \centering
    \includegraphics[width=0.95\linewidth]{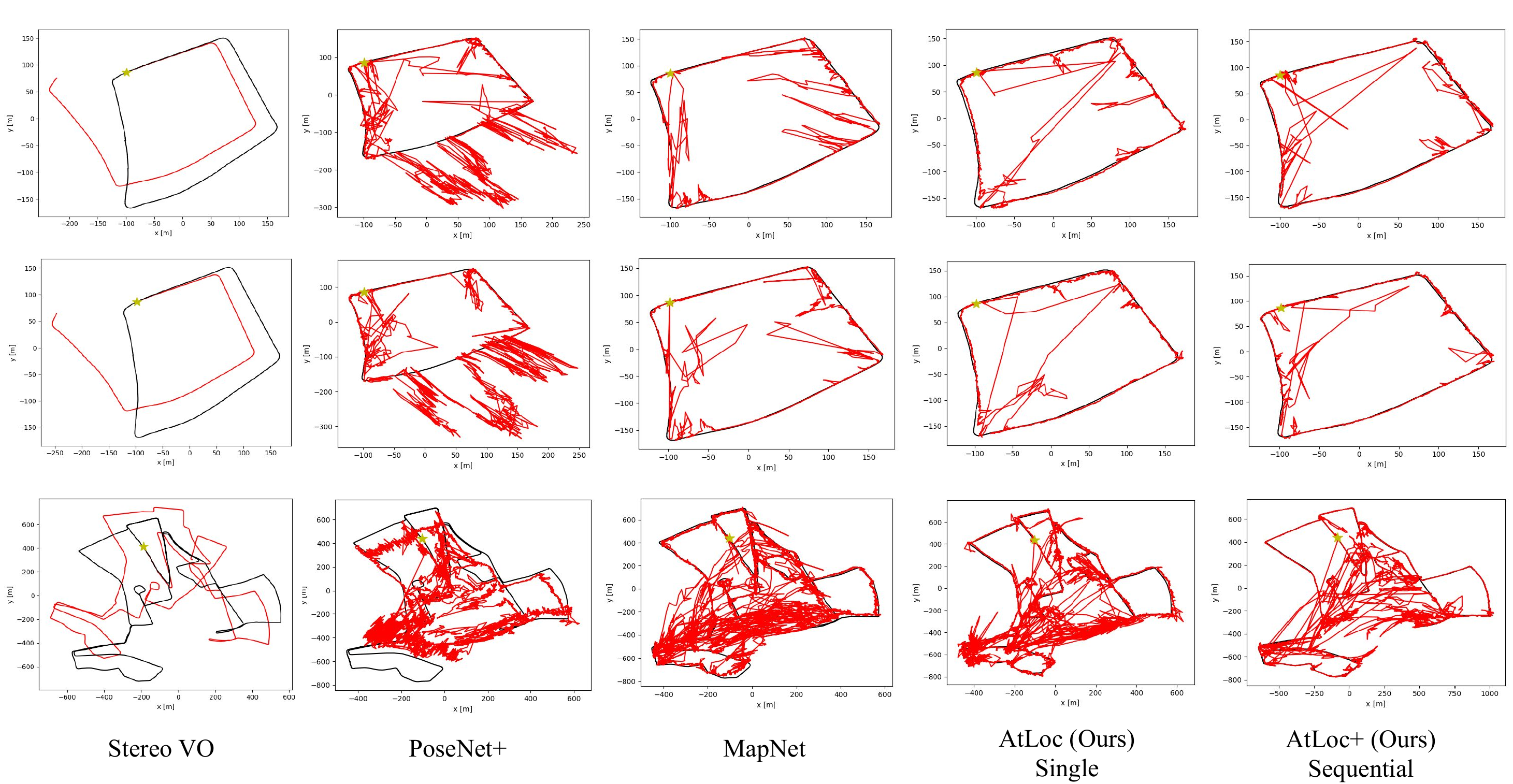}
    \caption{\textbf{Trajectories on LOOP1 (top), LOOP2 (middle) and FULL1 (bottom) of Oxford RobotCar.} The ground truth trajectories are shown in black lines while the red lines are the predictions. The star in the trajectory represents the starting point.}
    \label{fig:Trajectory}
    \vspace{-0.5cm}
\end{figure*}
\paragraph{Oxford RobotCar} \cite{maddern20171}
was recorded by an autonomous Nissan LEAF car in Oxford, UK over several periods for a year. This dataset exhibits substantial observations in the presence of various weather conditions, such as sunny and snowy days, as well as different lighting conditions, e.g., dim and glare roadworks. Moreover, we also found many dynamic or affordable objects in the scenes (e.g., parked/moving vehicles, cyclists and pedestrians), making this dataset particularly challenging for vision-based relocalization tasks. For a fair comparison, we follow the same evaluation strategy of MapNet \cite{brahmbhatt2018geometry,xue2019local} and use two subsets of this dataset in our experiments, labelled as LOOP and FULL (length-based) respectively. More details about these two sequences can be found in Table~\ref{tab:train_test_split}
In terms of implementation, we take the images recorded by the centre camera at a resolution of $1280\times960$ as the input to our network. The corresponding ground truth poses are labelled by the interpolations of INS measurements.

\paragraph{Baselines}
To validate the performance of our proposed network, we compare the results of several competing approaches. For experiments on 7 Scenes, we choose the following mainstream single-image-based methods: PoseNet \cite{kendall2015posenet}, Bayesian PoseNet \cite{kendall2016modelling}, PoseNet Spatial-LSTM \cite{walch2017image}, Hourglass\cite{melekhov2017image} and PoseNet17 \cite{kendall2017geometric}. Moreover, we also report the results of temporal approaches, VidLoc\cite{clark2017vidloc} and MapNet, for comparison. For the outdoor Oxford RobotCar dataset, Stereo VO \cite{maddern20171} and PoseNet+ (aka. ResNet34+log q) \cite{brahmbhatt2018geometry} are selected as our baselines. It is worth mentioning that PostNet+ is the best variant of PoseNet \cite{kendall2015posenet} on the RobotCar dataset \cite{brahmbhatt2018geometry}. Lastly, we also report the performance of MapNet\cite{brahmbhatt2018geometry}, the state-of-art method on this dataset using a sequence of images for relocalization. Note that as sequence based methods can exploit temporal constraints, they generally perform better than single-image based approaches. We nevertheless still compare with MapNet in evaluation to examine how accurate our single-image based AtLoc is.

\subsection{Experiments on 7 Scenes}
7Scenes Dataset contains 7 static indoor scenes with a large number of images captured in an office building. We take all scenes for comprehensive performance evaluation. 

\paragraph{Quantitative Results} Table~\ref{tbl:7scenes} and Table~\ref{tbl:7scenes_Seq} summarize the performance of all methods. Clearly, we can see that our method outperforms other single-image-based methods, with a $\mathbf{13\%}$ improvement in position accuracy and a $\mathbf{7\%}$ improvement in rotation than the best single-image based baseline. In particular, AtLoc achieves the best performance gain in large texture-less (such as whiteboard) and highly texture-repetitive (such as stairs) scenarios. AtLoc reduces the position error from $0.35m$ to $0.26m$ and the rotation error from $12.4\degree$ to $10.5\degree$ in the scene of $Stairs$, which is a significant improvement over prior arts. In other regular scenes, AtLoc still reaches a comparable accuracy against baselines. By using only a single image, AtLoc achieves a superior accuracy compared with MapNet, despite the uses of image sequences and handcrafted geometric constraints in the MapNet design. Last but not least, after incorporating temporal constraints, AtLoc+ further narrows the median position and rotation errors to $0.19m$ and $7.08\degree$ respectively, outperforming MapNet by a large margin.
\begin{figure*}[t]
    \centering
    \includegraphics[width=0.9\linewidth]{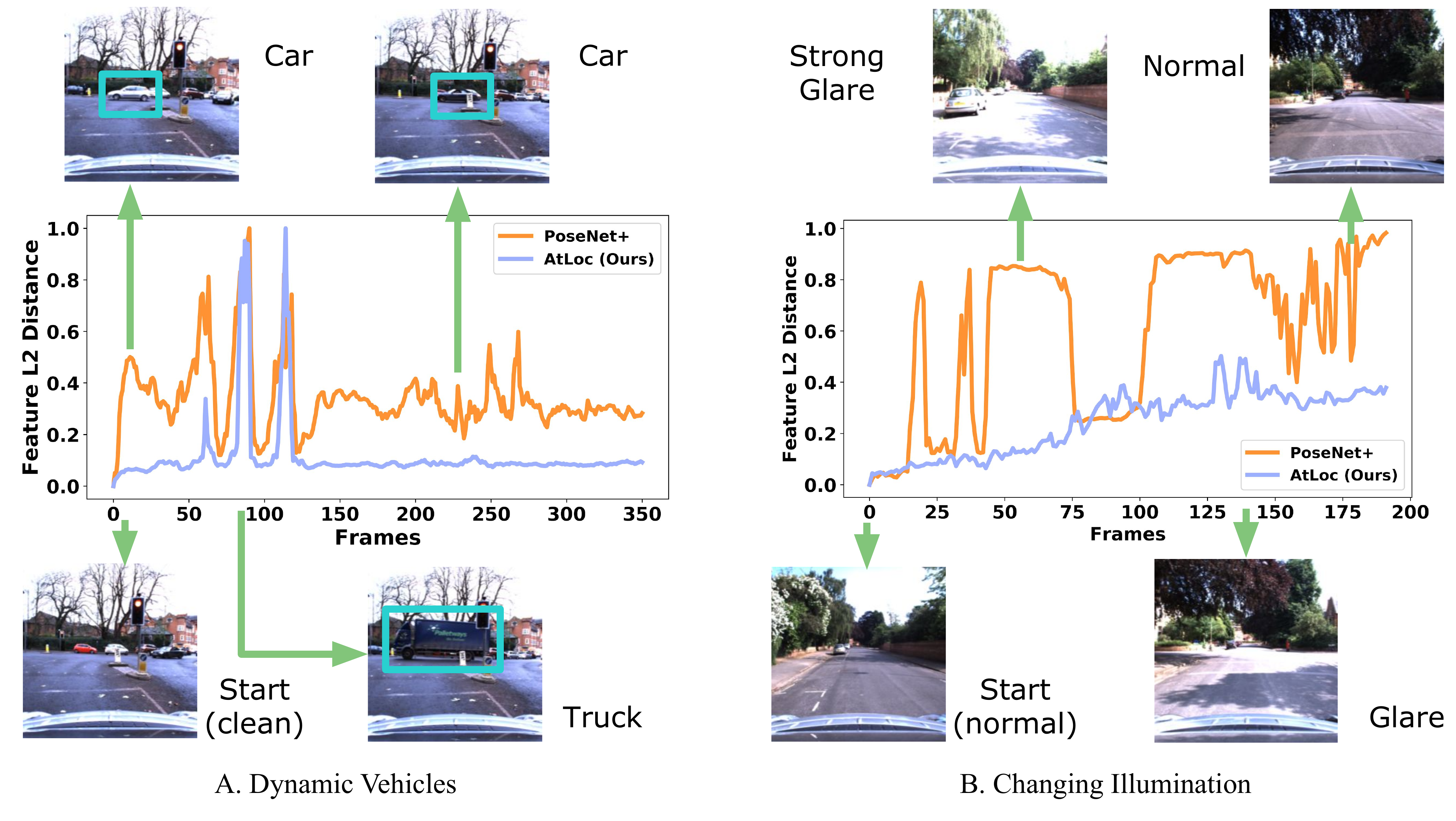}
    \vspace{-0.3cm}
    \caption{\textbf{Feature distance comparisons under different dynamic disturbances.} (a) Dynamic vehicles and (b) Changing illumination. Feature distances of AtLoc reasonably change with the motion status of the camera and are agnostic to various dynamics, while PoseNet suffers in both experiments.}
    \label{fig:feature_distance}
\vspace{-0.5cm}
\end{figure*}
\paragraph{Qualitative Results} To deeply understand the reasons behind these improvements, we visualize the attention maps of some scenes from 7Scenes. As shown in Figure~\ref{fig:chess_att}, by using attention, AtLoc focuses more on geometrically meaningful areas (e.g. key points and lines) rather than feature-less regions and shows better consistency over time. In contrast, the saliency maps of PoseNet are relatively scattered and tend to focus on random regions in the view. A video that compares the saliency map between PoseNet and AtLoc in detail can be found at \url{https://youtu.be/_xObJ1xwt94}.

\subsection{Localization Results on Oxford RobotCar}
We next evaluate our approach on Oxford RobotCar dataset. Due to the substantial dynamics over the long collection period, this dataset is very challenging and strictly demands high robustness and adaptability of a recloazalition model. 

\paragraph{Quantitative Results} Table~\ref{tbl:robotcar_results} shows the comparison of our methods against PoseNet+ and MapNet. Compared with PoseNet+, AtLoc presents significant improvements on both LOOP trajectories and FULL trajectories. The mean position accuracy is improved from $25.29m$ to $8.61m$ on LOOP1, and $28.81m$ to $8.86m$ on LOOP2.
The largest performance gains are observed on FULL1 and FULL2, where our approach outperforms PoseNet+ by $76.5\%$ and $63.3\%$. When compared against the sequence-based MapNet, our AtLoc has obvious accuracy gain on all cases. Even for the unfavorable routes (FULL1 and FULL2), AtLoc still provides $28.5\%$ and $30.3\%$ improvements over MapNet. Futhermore, AtLoc+ significantly improves the position accuracy to $19.7m$ and the rotation accuracy to $5.83\degree$ after temporal constraints introduced.
\vspace{-0.2cm}

\paragraph{Qualitative Results} 
We now investigate why AtLoc significantly outperforms baselines on the Oxford RoboCar dataset.
In Figure~\ref{fig:Trajectory}, we plot the predictions of LOOP1 (top), LOOP2 (middle) and FULL1 (bottom) by Stereo VO, Posenet+, MapNet, AtLoc and AtLoc+. Stereo VO is the official baseline from Oxford RobotCar. Although Stereo VO has very smooth predicted trajectories, it suffers from significant drifts as route length increases. Due to strong local similarity, there are many outliers predicted by PoseNet+. These outliers, however, are significantly reduced in by AtLoc and AtLoc+.
By looking into the saliency maps (Figure~\ref{fig:robotcar_att}), we found PoseNet+ heavily relies on texture-less regions, such as local road surface (top), dynamic cars (middle) and affordance objects such as bicycles (bottom). These regions are either too similar in appearance or unreliable due to changes overtime, making pose estimation difficult. By contrast, our attention-guided AtLoc is able to automatically focus on unique, static and stable areas/objects, including vanishing lines and points (top), buildings (middle and bottom). These areas are tightly related to the latent geometric features of an environment, enabling robust pose estimation in the wild. 

To further understand the efficacy of the attention mechanism, we depict the feature distances for a sequence of images. Specifically, we select a starting frame in the trajectory and then calculate feature distances ($L_2$) of subsequent frames to the starting frame. Features are extracted by PoseNet+ and AtLoc respectively, with the intention to understand to what extent the attention mechanism can help extract robust features. For experiments, we plot the distance profile under two cases: (i) dynamic vehicles and (ii) changing illumination. As we can see in Fig.\ref{fig:feature_distance} (left), when the camera is static (i.e., the data-collection car is not moving), PoseNet+ is sensitive to dynamic objects entering the scene, resulting in a large variation of distances. In contrary, thanks to the adopted attention mechanism, AtLoc is robust to these moving vehicles and provides more stable features overall. Distance spikes are only observed when a large truck enters/leaves the scene, in which a substantial portion of view is blocked/revealed to the camera.
On the right side of Fig.\ref{fig:feature_distance}, the extracted features of PoseNet+ suffer from illumination changes and gives abrupt shifts under different levels of glare. The features extracted by AtLoc, however, consistently change as the camera moves forward, agnostic to various lighting conditions.

\subsection{Ablation Study and Efficiency Evaluation}

\begin{table}[t]
\centering
\resizebox{\linewidth}{!}{
\begin{tabular}{l|cccc}
\multicolumn{1}{c|}{} & \multicolumn{4}{c}{7Scenes}                                                                                                                                     \\ \hline
Scene                 & \begin{tabular}[c]{@{}c@{}}AtLoc\\ Basic\end{tabular} & \begin{tabular}[c]{@{}c@{}}AtLoc\\ Basic+LSTM\end{tabular} & AtLoc                & AtLoc+               \\ \hline
Chess                 & 0.11m, 4.29                                           & 0.13m, 4.26                                                & 0.10m, 4.07          & \textbf{0.10m, 3.18} \\
Fire                  & 0.29m, 12.1                                           & 0.27m, 11.7                                                & 0.25m, 11.4          & \textbf{0.26m, 10.8} \\
Heads                 & 0.19m, 12.2                                           & 0.16m, 12.3                                                & 0.16m, 11.8          & \textbf{0.14m, 11.4} \\
Office                & 0.19m, 6.35                                           & 0.20m, 5.74                                                & 0.17m, 5.34          & \textbf{0.17m, 5.16} \\
Pumpkin               & 0.22m, 5.05                                           & 0.26m, 4.19                                                & 0.21m, 4.37          & \textbf{0.20m, 3.94} \\
Kitchen               & 0.25m,5.27                                            & 0.19m, \textbf{4.63}                                       & 0.23m, 5.42          & \textbf{0.16m, 4.90} \\
Stairs                & 0.30m, 11.3                                           & 0.29m, 12.1                                                & \textbf{0.26m}, 10.5 & \textbf{0.29m, 10.2} \\ \hline
Average               & 0.22m, 8.07                                           & 0.21m, 7.83                                                & 0.20m, 7.56          & \textbf{0.19m, 7.08} \\ \hline
\multicolumn{1}{c|}{} & \multicolumn{4}{c}{RobotCar}                                                                                                                                    \\ \hline
LOOP1                 & 25.29m, 17.45                                         & 29.71m, 15.72                                              & 8.61m, 4.58          & \textbf{7.82m, 3.62} \\
LOOP2                 & 28.81m, 19.62                                         & 32.79m, 17.76                                              & 8.86m, 4.67          & \textbf{7.24m, 3.60} \\
FULL1                 & 125.6m, 27.10                                         & 48.29m, 17.18                                              & 29.6m, 12.4          & \textbf{21.0m, 6.15} \\
FULL2                 & 131.1m, 26.05                                         & 67.62m, 11.40                                              & 48.2m, 11.1          & \textbf{42.6m, 9.95} \\ \hline
Average               & 77.70m, 22.56                                         & 44.60m, 15.52                                              & 23.8m, 8.19          & \textbf{19.7m, 5.83}
\end{tabular}
}
\caption{\textbf{Ablation study of AtLoc on 7 Scenes and Oxford RobotCar.} AtLoc (Basic) denotes the model without using attention. }
	\label{tbl:ablation}
    \vspace{-0.5cm}
\end{table}

We conduct an ablation study on the introduced attention module above 7 scenes and Oxford RoboCar datasets.
In Table~\ref{tbl:ablation} , AtLoc is compared with a basic version without the attention module, a LSTM version by replacing the attention module with the LSTM module, and a sequence enhanced version with temporal constraints. The rest modules are kept as the same for a fair comparison.
The comparison of AtLoc (Basic) and AtLoc indicates that the model performance clearly increases on both datasets by adopting the self-attention into the pose regression model: it shows a $9\%$ improvement in location accuracy and $6\%$ in rotation accuracy on the 7 Scenes dataset; AtLoc achieves an average localization accuracy of $23.8m$ and an average rotation accuracy of only $8.19\degree$ on Oxford RobotCar dataset. The LSTM based AtLoc is only better than Basic AtLoc. With temporal constraints, AtLoc+ yields the best performance, obtaining a median error of $0.19m$, $7.08\degree$ on 7 Scenes, and $19.7m$, $5.83\degree$ on RobotCar.

To evaluate the efficiency of our proposed AtLoc, we analyze the average running time of three models - MapNet, PoseLSTM and AtLoc. 
Among the three models, MapNet consumes the longest running time of $9.4 ms$ per frame, as it needs to process additional data from other sensory inputs and a sequence of images to apply geometric constraints. 
Due to the time-consuming recursive operations in LSTMs, PoseLSTM takes a running time of $9.2 ms$ per frame, $3.7 ms$ higher than its corresponding basic model PoseNet. 
In contrast, our proposed AtLoc achieves an ideal balance between the computational efficiency and localization accuracy, consuming only $6.3 ms$ per frame while obtaining the best localization performance.

\section{Conclusion and Discussion}
Camera localization is a challenging task in computer vision due to scene dynamics and high variability of environment appearance.
In this work, we presented a novel study of self-attention guided camera localization from a single image. 
The introduced self-attention can encourage the framework to learn geometrically robust features, mitigating the impacts from dynamic objects and changing illumination.
We demonstrate state-of-the-art results, even surpassing sequential based techniques in challenging scenarios. 
Further work includes refining the attention module and determining whether it can improve multi-frame camera pose regression.

\FloatBarrier
\bibliographystyle{aaai}

\bibliography{refs.bib}

\begin{thebibliography}{}

\bibitem[\protect\citeauthoryear{Arandjelovic \bgroup et al\mbox.\egroup
  }{2016}]{arandjelovic2016netvlad}
Arandjelovic, R.; Gronat, P.; Torii, A.; Pajdla, T.; and Sivic, J.
\newblock 2016.
\newblock Netvlad: Cnn architecture for weakly supervised place recognition.
\newblock In {\em CVPR}.

\bibitem[\protect\citeauthoryear{Bahdanau, Cho, and
  Bengio}{2014}]{bahdanau2014neural}
Bahdanau, D.; Cho, K.; and Bengio, Y.
\newblock 2014.
\newblock Neural machine translation by jointly learning to align and
  translate.
\newblock In {\em ICLR}.

\bibitem[\protect\citeauthoryear{Brachmann \bgroup et al\mbox.\egroup
  }{2017}]{brachmann2017dsac}
Brachmann, E.; Krull, A.; Nowozin, S.; Shotton, J.; Michel, F.; Gumhold, S.;
  and Rother, C.
\newblock 2017.
\newblock Dsac-differentiable ransac for camera localization.
\newblock In {\em CVPR}.

\bibitem[\protect\citeauthoryear{Brahmbhatt \bgroup et al\mbox.\egroup
  }{2018}]{brahmbhatt2018geometry}
Brahmbhatt, S.; Gu, J.; Kim, K.; Hays, J.; and Kautz, J.
\newblock 2018.
\newblock Geometry-aware learning of maps for camera localization.
\newblock In {\em CVPR}.

\bibitem[\protect\citeauthoryear{Cai, Shen, and Reid}{2018}]{cai2018hybrid}
Cai, M.; Shen, C.; and Reid, I.~D.
\newblock 2018.
\newblock A hybrid probabilistic model for camera relocalization.
\newblock In {\em BMVC}.

\bibitem[\protect\citeauthoryear{Chen \bgroup et al\mbox.\egroup
  }{2011}]{chen2011city}
Chen, D.~M.; Baatz, G.; K{\"o}ser, K.; Tsai, S.~S.; Vedantham, R.;
  Pylv{\"a}n{\"a}inen, T.; Roimela, K.; Chen, X.; Bach, J.; Pollefeys, M.;
  et~al.
\newblock 2011.
\newblock City-scale landmark identification on mobile devices.
\newblock In {\em CVPR}.

\bibitem[\protect\citeauthoryear{Chen \bgroup et al\mbox.\egroup
  }{2019}]{chen2019selective}
Chen, C.; Rosa, S.; Miao, Y.; Lu, C.~X.; Wu, W.; Markham, A.; and Trigoni, N.
\newblock 2019.
\newblock Selective sensor fusion for neural visual-inertial odometry.
\newblock In {\em CVPR}.

\bibitem[\protect\citeauthoryear{Cheng, Dong, and Lapata}{2016}]{cheng2016long}
Cheng, J.; Dong, L.; and Lapata, M.
\newblock 2016.
\newblock Long short-term memory-networks for machine reading.
\newblock In {\em EMNLP}.

\bibitem[\protect\citeauthoryear{Clark \bgroup et al\mbox.\egroup
  }{2017}]{clark2017vidloc}
Clark, R.; Wang, S.; Markham, A.; Trigoni, N.; and Wen, H.
\newblock 2017.
\newblock Vidloc: A deep spatio-temporal model for 6-dof video-clip
  relocalization.
\newblock In {\em CVPR}.

\bibitem[\protect\citeauthoryear{{Deng} \bgroup et al\mbox.\egroup
  }{2009}]{imagenet}
{Deng}, J.; {Dong}, W.; {Socher}, R.; {Li}, L.; {Kai Li}; and {Li Fei-Fei}.
\newblock 2009.
\newblock Imagenet: A large-scale hierarchical image database.
\newblock In {\em CVPR}.

\bibitem[\protect\citeauthoryear{Dou \bgroup et al\mbox.\egroup
  }{2019}]{dou2019dynamic}
Dou, Z.-Y.; Tu, Z.; Wang, X.; Wang, L.; Shi, S.; and Zhang, T.
\newblock 2019.
\newblock Dynamic layer aggregation for neural machine translation with
  routing-by-agreement.
\newblock In {\em AAAI}.

\bibitem[\protect\citeauthoryear{Fu \bgroup et al\mbox.\egroup
  }{2019}]{fu2019sta}
Fu, Y.; Wang, X.; Wei, Y.; and Huang, T.
\newblock 2019.
\newblock Sta: Spatial-temporal attention for large-scale video-based person
  re-identification.
\newblock In {\em AAAI}.

\bibitem[\protect\citeauthoryear{He \bgroup et al\mbox.\egroup
  }{2016}]{he2016deep}
He, K.; Zhang, X.; Ren, S.; and Sun, J.
\newblock 2016.
\newblock Deep residual learning for image recognition.
\newblock In {\em CVPR}.

\bibitem[\protect\citeauthoryear{Kendall and
  Cipolla}{2016}]{kendall2016modelling}
Kendall, A., and Cipolla, R.
\newblock 2016.
\newblock Modelling uncertainty in deep learning for camera relocalization.
\newblock In {\em ICRA}.

\bibitem[\protect\citeauthoryear{Kendall and
  Cipolla}{2017}]{kendall2017geometric}
Kendall, A., and Cipolla, R.
\newblock 2017.
\newblock Geometric loss functions for camera pose regression with deep
  learning.
\newblock In {\em CVPR}.

\bibitem[\protect\citeauthoryear{Kendall, Grimes, and
  Cipolla}{2015}]{kendall2015posenet}
Kendall, A.; Grimes, M.; and Cipolla, R.
\newblock 2015.
\newblock Posenet: A convolutional network for real-time 6-dof camera
  relocalization.
\newblock In {\em CVPR}.

\bibitem[\protect\citeauthoryear{Kingma and Ba}{2014}]{kingma2014adam}
Kingma, D.~P., and Ba, J.
\newblock 2014.
\newblock Adam: A method for stochastic optimization.
\newblock In {\em ICLR}.

\bibitem[\protect\citeauthoryear{Kingma and Dhariwal}{2018}]{kingma2018glow}
Kingma, D.~P., and Dhariwal, P.
\newblock 2018.
\newblock Glow: Generative flow with invertible 1x1 convolutions.
\newblock In {\em NIPS}.

\bibitem[\protect\citeauthoryear{Liu \bgroup et al\mbox.\egroup
  }{2019}]{liu2019perceptual}
Liu, A.; Liu, X.; Fan, J.; Ma, Y.; Zhang, A.; Xie, H.; and Tao, D.
\newblock 2019.
\newblock Perceptual-sensitive gan for generating adversarial patches.
\newblock In {\em AAAI}.

\bibitem[\protect\citeauthoryear{Liu, Li, and Dai}{2017}]{liu2017efficient}
Liu, L.; Li, H.; and Dai, Y.
\newblock 2017.
\newblock Efficient global 2d-3d matching for camera localization in a
  large-scale 3d map.
\newblock In {\em ICCV}.

\bibitem[\protect\citeauthoryear{Maddern \bgroup et al\mbox.\egroup
  }{2017}]{maddern20171}
Maddern, W.; Pascoe, G.; Linegar, C.; and Newman, P.
\newblock 2017.
\newblock 1 year, 1000 km: The oxford robotcar dataset.
\newblock In {\em The International Journal of Robotics Research}, volume~36,
  3--15.

\bibitem[\protect\citeauthoryear{Melekhov \bgroup et al\mbox.\egroup
  }{2017}]{melekhov2017image}
Melekhov, I.; Ylioinas, J.; Kannala, J.; and Rahtu, E.
\newblock 2017.
\newblock Image-based localization using hourglass networks.
\newblock In {\em ICCV}.

\bibitem[\protect\citeauthoryear{Parisotto \bgroup et al\mbox.\egroup
  }{2018}]{parisotto2018global}
Parisotto, E.; Singh~Chaplot, D.; Zhang, J.; and Salakhutdinov, R.
\newblock 2018.
\newblock Global pose estimation with an attention-based recurrent network.
\newblock In {\em CVPR Workshops}.

\bibitem[\protect\citeauthoryear{Parmar \bgroup et al\mbox.\egroup
  }{2018}]{parmar2018image}
Parmar, N.; Vaswani, A.; Uszkoreit, J.; Kaiser, {\L}.; Shazeer, N.; Ku, A.; and
  Tran, D.
\newblock 2018.
\newblock Image transformer.
\newblock In {\em ICML}.

\bibitem[\protect\citeauthoryear{Purkait, Zhao, and
  Zach}{2018}]{purkait2018synthetic}
Purkait, P.; Zhao, C.; and Zach, C.
\newblock 2018.
\newblock Synthetic view generation for absolute pose regression and image
  synthesis.
\newblock In {\em BMVC}.

\bibitem[\protect\citeauthoryear{Sattler \bgroup et al\mbox.\egroup
  }{2019}]{sattler2019understanding}
Sattler, T.; Zhou, Q.; Pollefeys, M.; and Leal-Taixe, L.
\newblock 2019.
\newblock Understanding the limitations of cnn-based absolute camera pose
  regression.
\newblock In {\em CVPR}.

\bibitem[\protect\citeauthoryear{Sattler, Leibe, and
  Kobbelt}{2012}]{sattler2012improving}
Sattler, T.; Leibe, B.; and Kobbelt, L.
\newblock 2012.
\newblock Improving image-based localization by active correspondence search.
\newblock In {\em ECCV}.

\bibitem[\protect\citeauthoryear{Shotton \bgroup et al\mbox.\egroup
  }{2013}]{shotton2013scene}
Shotton, J.; Glocker, B.; Zach, C.; Izadi, S.; Criminisi, A.; and Fitzgibbon,
  A.
\newblock 2013.
\newblock Scene coordinate regression forests for camera relocalization in
  rgb-d images.
\newblock In {\em CVPR},  2930--2937.

\bibitem[\protect\citeauthoryear{Szegedy \bgroup et al\mbox.\egroup
  }{2015}]{szegedy2015going}
Szegedy, C.; Liu, W.; Jia, Y.; Sermanet, P.; Reed, S.; Anguelov, D.; Erhan, D.;
  Vanhoucke, V.; and Rabinovich, A.
\newblock 2015.
\newblock Going deeper with convolutions.
\newblock In {\em CVPR}.

\bibitem[\protect\citeauthoryear{Vaswani \bgroup et al\mbox.\egroup
  }{2017}]{vaswani2017attention}
Vaswani, A.; Shazeer, N.; Parmar, N.; Uszkoreit, J.; Jones, L.; Gomez, A.~N.;
  Kaiser, {\L}.; and Polosukhin, I.
\newblock 2017.
\newblock Attention is all you need.
\newblock In {\em NIPS}.

\bibitem[\protect\citeauthoryear{Walch \bgroup et al\mbox.\egroup
  }{2017}]{walch2017image}
Walch, F.; Hazirbas, C.; Leal-Taixe, L.; Sattler, T.; Hilsenbeck, S.; and
  Cremers, D.
\newblock 2017.
\newblock Image-based localization using lstms for structured feature
  correlation.
\newblock In {\em CVPR}.

\bibitem[\protect\citeauthoryear{Wang \bgroup et al\mbox.\egroup
  }{2018a}]{wang2018end}
Wang, S.; Clark, R.; Wen, H.; and Trigoni, N.
\newblock 2018a.
\newblock End-to-end, sequence-to-sequence probabilistic visual odometry
  through deep neural networks.
\newblock In {\em The International Journal of Robotics Research}, volume~37,
  513--542.

\bibitem[\protect\citeauthoryear{Wang \bgroup et al\mbox.\egroup
  }{2018b}]{wang2018non}
Wang, X.; Girshick, R.; Gupta, A.; and He, K.
\newblock 2018b.
\newblock Non-local neural networks.
\newblock In {\em CVPR}.

\bibitem[\protect\citeauthoryear{Xu \bgroup et al\mbox.\egroup
  }{2015}]{xu2015show}
Xu, K.; Ba, J.; Kiros, R.; Cho, K.; Courville, A.; Salakhudinov, R.; Zemel, R.;
  and Bengio, Y.
\newblock 2015.
\newblock Show, attend and tell: Neural image caption generation with visual
  attention.
\newblock In {\em ICML}.

\bibitem[\protect\citeauthoryear{Xue \bgroup et al\mbox.\egroup
  }{2019}]{xue2019local}
Xue, F.; Wang, X.; Yan, Z.; Wang, Q.; Wang, J.; and Zha, H.
\newblock 2019.
\newblock Local supports global: Deep camera relocalization with sequence
  enhancement.
\newblock In {\em ICCV}.

\bibitem[\protect\citeauthoryear{Yang \bgroup et al\mbox.\egroup
  }{2019}]{yang2019context}
Yang, B.; Li, J.; Wong, D.~F.; Chao, L.~S.; Wang, X.; and Tu, Z.
\newblock 2019.
\newblock Context-aware self-attention networks.
\newblock In {\em AAAI}.

\bibitem[\protect\citeauthoryear{Yuan, Mei, and Zhu}{2019}]{yuan2019find}
Yuan, Y.; Mei, T.; and Zhu, W.
\newblock 2019.
\newblock To find where you talk: Temporal sentence localization in video with
  attention based location regression.
\newblock In {\em AAAI}.

\bibitem[\protect\citeauthoryear{Zhang \bgroup et al\mbox.\egroup
  }{2018}]{zhang2018self}
Zhang, H.; Goodfellow, I.; Metaxas, D.; and Odena, A.
\newblock 2018.
\newblock Self-attention generative adversarial networks.
\newblock In {\em ICML}.

\bibitem[\protect\citeauthoryear{Zhaoyang~Huang}{2019}]{huang2019prior}
Zhaoyang~Huang, Yan~Xu, J. S. X. Z. H. B. G.~Z.
\newblock 2019.
\newblock Prior guided dropout for robust visual localization in dynamic
  environments.
\newblock In {\em ICCV}.

\end{thebibliography}

\end{document}